\pdfoutput=1

\documentclass[11pt]{article}

\usepackage[final]{acl}

\usepackage{times}
\usepackage{latexsym}

\usepackage[T1]{fontenc}

\usepackage[utf8]{inputenc}

\usepackage{microtype}

\usepackage{inconsolata}

\usepackage{graphicx}

\usepackage{microtype}

\usepackage{amssymb,bm,mathtools,color}

\usepackage[normalem]{ulem}
\usepackage{enumitem}

\usepackage{booktabs}
\usepackage{multirow}
\usepackage{tabularx}
\usepackage{multicol}
\usepackage{multirow}
\usepackage{colortbl}
\usepackage{hhline}
\usepackage{stfloats}

\newcolumntype{Y}{>{\centering\arraybackslash}X}

\usepackage{xcolor}
\definecolor{red}{rgb}{0.74,0.08,0.10}
\definecolor{green}{rgb}{0.26,0.49,0.18}
\definecolor{blue}{rgb}{0.22,0.53,0.75}

\usepackage{colortbl}
\definecolor{Gray}{gray}{0.9}
\definecolor{LightCyan}{rgb}{0.75,1,1}

\makeatletter
\newcommand\notsotiny{\@setfontsize\notsotiny\@viiipt\@ixpt}
\makeatother
\usepackage{booktabs}
\usepackage{multirow}
\usepackage{graphicx}

\usepackage{pifont}
\usepackage{amsmath}
\usepackage{amssymb}
\usepackage{ifthen}
\usepackage{dsfont}
\usepackage{bold-extra}
\usepackage{courier}

\usepackage{listings}
\usepackage{xcolor}
\usepackage{ltablex}
\usepackage{longtable}
\usepackage{multicol}

\usepackage{tikz}
\usetikzlibrary{bayesnet}
\usepackage[linguistics]{forest}

\usepackage{caption}
\usepackage{cancel}

\usepackage{xspace}
\usepackage{comment}
\usepackage{color,soul}

\usepackage{float}

\usepackage[noabbrev,capitalize]{cleveref} 

\crefname{page}{page}{pages}
\crefname{footnote}{footnote}{footnotes}   
\crefname{equation}{equation}{equations}   

\creflabelformat{equation}{#2\textup{#1}#3}

\crefname{line}{line}{lines}               
\crefrangeformat{line}{lines #3#1#4--#5#2#6}
\crefname{lstlsting}{Listing}{Listings}   
\crefname{section}{\S}{\S\S}
\Crefname{section}{\S}{\S\S}    
\crefformat{section}{#2\S#1#3}  
\Crefformat{section}{#2\S#1#3}
\crefrangeformat{section}{\S\S#3#1#4--#5#2#6}
\Crefrangeformat{section}{\S\S#3#1#4--#5#2#6}
\crefmultiformat{section}{\S#2#1#3}{ and~\S#2#1#3}{, \S#2#1#3}{ and~\S#2#1#3}
\Crefmultiformat{section}{\S#2#1#3}{ and~\S#2#1#3}{, \S#2#1#3}{ and~\S#2#1#3}
\crefrangemultiformat{section}{\S\S#3#1#4--#5#2#6}{ and~\S\S#3#1#4--#5#2#6}{, \S\S#3#1#4--#5#2#6}{ and~\S\S#3#1#4--#5#2#6}
\Crefrangemultiformat{section}{\S\S#3#1#4--#5#2#6}{ and~\S\S#3#1#4--#5#2#6}{, \S\S#3#1#4--#5#2#6}{ and~\S\S#3#1#4--#5#2#6}

\usepackage{makecell}

\usepackage[most]{tcolorbox}

\newcommand{\prob}[2][]{p\ifthenelse{\not\equal{}{#1}}{_{#1}}{}(#2)} 
\newcommand{\expect}[2][]{\text{\bf E}\ifthenelse{\not\equal{}{#1}}{_{#1}}{}\!\left[#2\right]}
\newcommand{\var}[2][]{\text{\bf Var}\ifthenelse{\not\equal{}{#1}}{_{#1}}{}\!\left[#2\right]}

\newcommand{\ie}{\emph{i.e.}, }
\newcommand{\eg}{\emph{e.g.}, }

\newcommand{\zsteer}{$z_{steer}$}

\newcommand{\acuteprotocol}{\textsc{acute}}
\newcommand{\euro}{\textsc{euro}}
\newcommand{\mice}{\textsc{mice}}
\newcommand{\kernelReg}{NWKR}

\newcommand{\makename}[3][s]{%
  \expandafter\newcommand\csname #2\endcsname{#3\xspace}%
  \expandafter\newcommand\csname #2s\endcsname{#3#1\xspace}%
}

\makename{methodNameAbbr}{\textsc{acute}}
\makename{euroauc}{\textsc{auc-euro}}
\makename{aucmse}{\texttt{AUC-MSE}}
\makename{scauc}{\texttt{SCAUC}}

\newtcolorbox{keyquestion}[2][]{
  colback=blue!5!white,    
  colframe=blue!75!black,  
  fonttitle=\bfseries,     
  title=#2,      
  arc=4pt,                 
  boxrule=1pt,             
  #1                       
}

\newcommand{\cmulogo}[2]{%
  \raisebox{#1}{\includegraphics[height=0.8em]{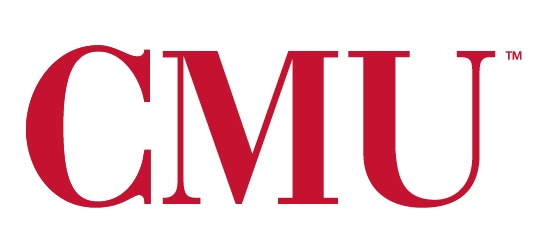}}%
  \hspace{0.1em}%
  \raisebox{#2}{\includegraphics[height=0.61em]{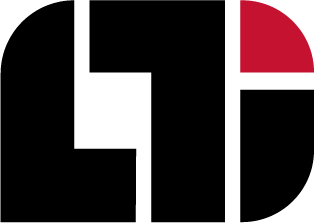}}%
}

\title{Harnessing the Latent Space:\\From Steering Vectors to Model Calibrators for Control and Trust}
\author{\
Nishant Subramani \cmulogo{0.3ex}{ 0.45ex}
\ \ \ \ \ 
\\
\cmulogo{-0.2ex}{-0.05ex}: Carnegie Mellon University, Language Technologies Institute \\
$\texttt{{nishant2}@cs.cmu.edu}$ \\
}

\begin{document}
\maketitle
\frenchspacing

\begin{abstract}
Language models have changed from unreliable text generators to highly-capable large models with trillions of parameters.
Capability increases come hand-in-hand with increases in scale, making understanding the internal representations of models more challenging.
Since millions of users increasing rely on language models to interact with external tools or make decisions in medium or high-stakes scenarios, we need to establish control over model behavior and know when to trust model outputs.
In this paper, we discuss our contributions on harnessing the latent spaces by proposing steering vectors for \emph{control} and developing latent space-based model calibrators for \emph{trust}.
Together, our contributions help demystify the latent spaces of language models and offer new insights into how to harness model internals to build more trustworthy language technology.
\end{abstract}
\section{Introduction}
\label{sec: introduction}

Neural network language models (LMs) have evolved from small, unreliable text generators to very large models capable of solving complex reasoning tasks~\citep[\emph{inter alia}]{peters-etal-2018-deep, radford2019language, groeneveld-etal-2024-olmo, yang2025qwen3, team2025gemma}.
Despite the vast capability increases, analyzing the internal representations of trillion-parameter models is challenging.
Due to this, the NLP community has increasingly treated models as black boxes, neglecting understanding the inner-workings of models.
Even though large language models (LLMs) are scaled to millions of users, increasingly interact with external tools~\citep{Qu2024ToolLW}, and make decisions in medium and high-stakes scenarios~\citep{Thirunavukarasu2023LargeLM}, we rely by-and-large on simple behavioral observation~\citep[\emph{inter alia}]{Hendrycks2020MeasuringMM, Srivastava2022BeyondTI, Liang2023HolisticEO}.
As a community, we must build fundamental understanding of the inner-workings of models and operationalize the internal representations of LLMs.
We need to establish \textbf{control} over model behavior to ensure safety and alignment and establish confidence estimation mechanisms which can accurately adjudicate \textbf{trust}.

We present four threads of research aimed to demystify and harness the latent spaces of language models.
To achieve model control, we show that LSTM-based language models can be minimally steered for exact generation (\cref{sec: steering_vectors_lstms}).
We then adapt to transformer-based models in~\cref{sec: steering_vectors_transformers}, showing both fine-grained and coarse-grained control via exact and concept-based steering.
Shifting to trustworthiness, we build model-internal confidence estimators (\mice) to calibrate LLM generations in tool-calling scenarios (\cref{sec: trust_mice}).
Lastly, we broaden the framework to new model families and tasks by proposing activation-based confidence, utility, and trust estimators (\acuteprotocol;~\cref{sec: trust_acute_euro}).
Together, our contributions offer actionable recipes to harness model internals to build more controllable, well-calibrated, and trustworthy language technologies.

\begin{figure}
    \centering
    \includegraphics[width=\linewidth]
    {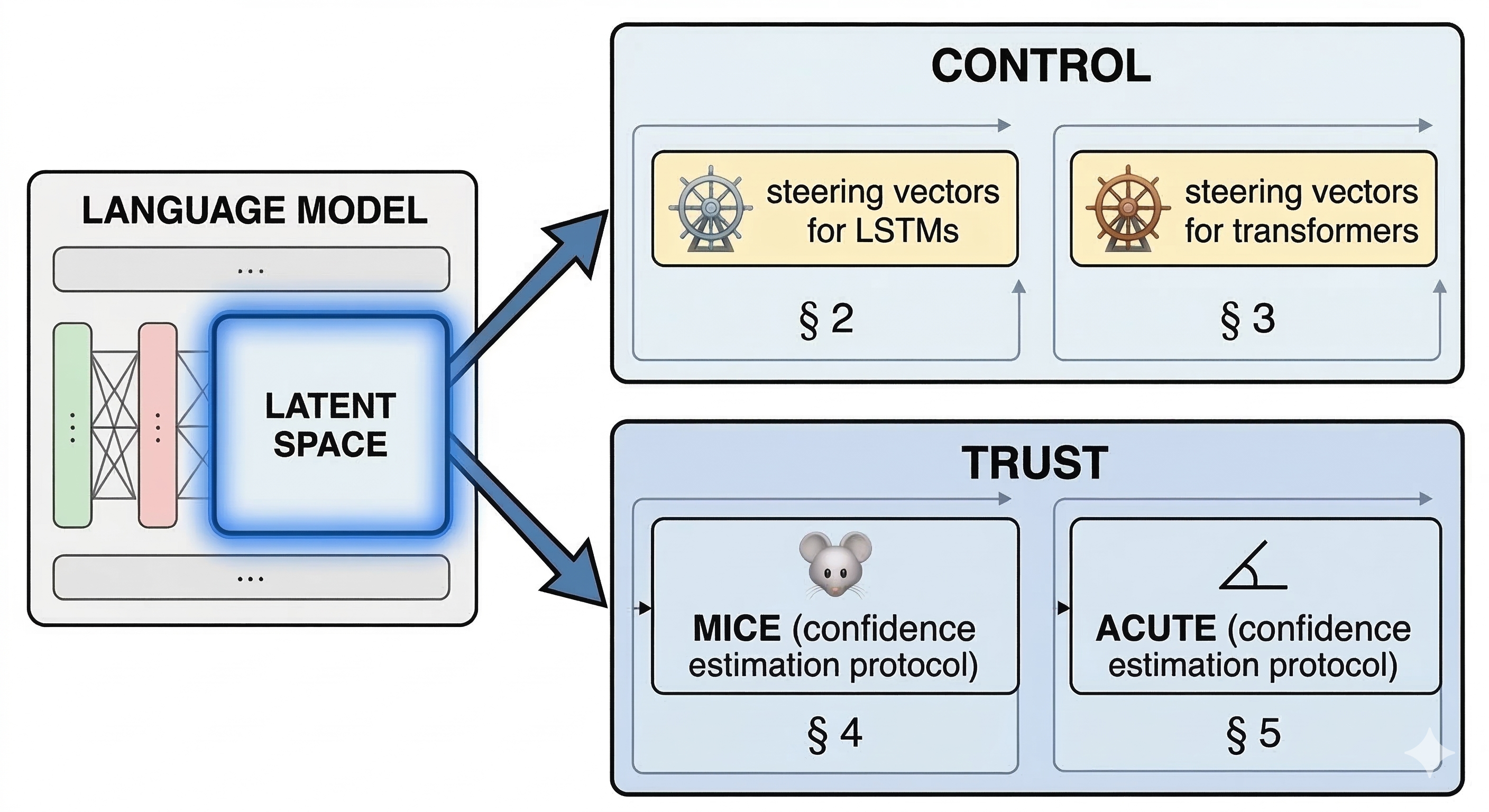}
    \caption{Our contributions on harnessing the latent spaces of language models:~\cref{sec: steering_vectors_lstms} and~\cref{sec: steering_vectors_transformers} focus on control, proposing steering vectors for the first time for LSTMs and transformer-based models.~\cref{sec: trust_mice} and~\cref{sec: trust_acute_euro} focus on trust, building model-internal confidence estimators to assess confidence of language model output generations.}
    \label{fig:process_fig}
\end{figure}
\section{Control: Steering Vectors for LSTMs~\citep{subramani2019can}}
\label{sec: steering_vectors_lstms}
We focus on control, specifically trying to answer one key question: 

\begin{keyquestion}{Key Question 1}
Can LSTM-based language models be steered to generate a desired sequence exactly without updating a single parameter?
\end{keyquestion}

\begin{figure*}[t]
    \centering
    \includegraphics[width=0.45\textwidth]{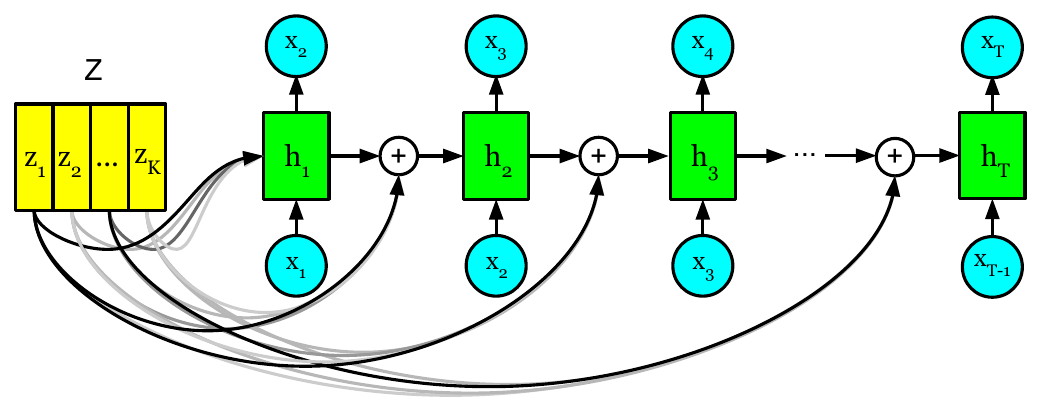}
    \hfill
    \includegraphics[width=0.53\textwidth]{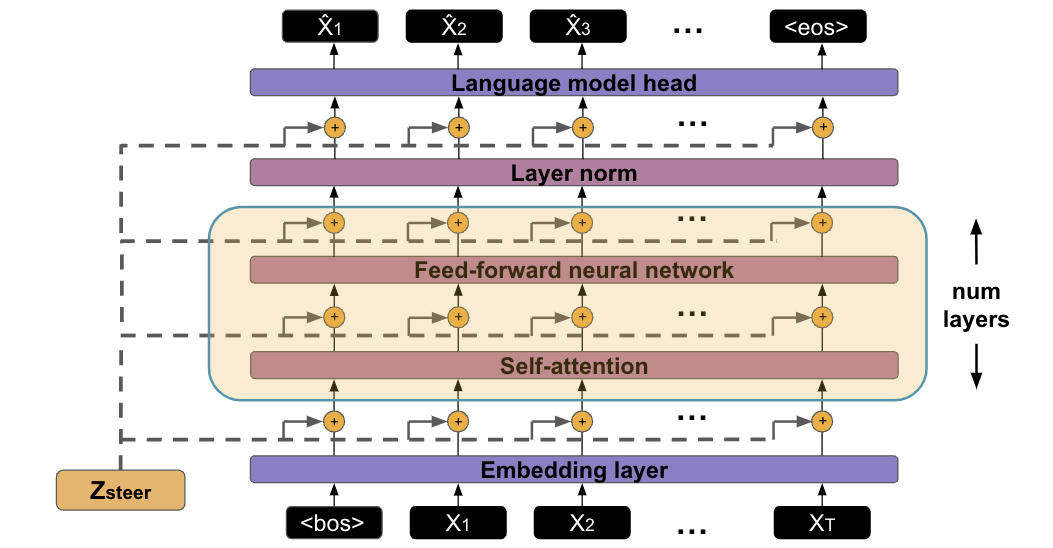}
    
    \caption{Here we show how the steering vector~\zsteer~can be injected into an LSTM-based language model (left) and a transformer-based one (right). On the left, Z is shown to have a larger dimension than the model dimension. If dim($Z$) equals the model dimension, $K=1$, and thus there is just one vector $z_1$.}
    \label{fig:steering_vec_process_fig}
\end{figure*}

\subsection{Prior Work}
In 2018, the transformer architecture proposed in~\citet{Vaswani2017AttentionIA} had yet to fully permeate the language model landscape and long short-term memory models (LSTMs;~\citealp{hochreiter1997long}) were still the predominant architecture for language modeling.
These language models were unreliable text generators. 
However, they have the potential to learn useful representations, so LMs started to be seen as general-purpose encoders~\citep[\emph{inter alia}]{dai2015semi, peters-etal-2018-deep, devlin-etal-2019-bert}.
Precise control of language model output, on the other hand, remained far out of reach, primarily due to the low quality of the underlying language models of the time.

\subsection{Our Contributions}
We explore whether language models could be used as \textit{general-purpose decoders}, something that we now take for granted, but at the time was an unknown.
For a pretrained language model to be used a general-purpose decoder, we need (1) to find a continuous-valued sentence representation (a steering vector) that can be fed into the frozen language model, (2) an encoder, likely task-specific, that can convert task inputs into steering vectors, and (3) for those steering vectors to \emph{causally} generate the desired output.
At the time, no work had shown this was possible, but now we take this for granted with advances in prompting and decoder-only LLMs.
In our work, we explore the possibilities of this in LSTM-based models, before prompting became popular.
Specifically, we ask whether LSTM-based models can be steered to generate a desired sequence exactly while keeping the underlying language model frozen.

\paragraph{Background}
To ask this, we first define the \textit{sentence space} of a recurrent language model.
Since the recurrent transition function $f_\theta = \mathbb{R}^{d} \times V \rightarrow \mathbb{R}^{d}$ defines a dynamical system based on the observations of tokens in a sequence.
As a result, the language model embeds a sequence of length $T$ as a $T+1$ step trajectory in a $d$-dimensional space, where $d$ is the dimension of the hidden state of the recurrent LM.
Next, we parametrize the sentence space into a flat-vector space $\mathcal{Z} \in \mathbb{R}^{d'}$ to better understand the sentence space of the LM.

To map the trajectory of hidden states to a flat vector in $\mathcal{Z}$, we add a bias term~\zsteer~$\in \mathcal{Z}$ to the previous hidden and cell state at each time step in the model and optimize~\zsteer~to maximize the log-probability of a given sequence.
Since we're adding~\zsteer~to every hidden and cell state as well as at every timestep, information contained in~\zsteer~will not degrade as quickly as if we just intervened at one location at one timestep.
Using this formulation, we can go back and forth, from vectors to sequences and vice-versa, and thus design experiments to test whether a frozen model can be steered to generate any sequence of interest.

To map from sequences to steering vectors (forward estimation), we modify the recurrent transition function, see~\cref{fig:steering_vec_process_fig} for details:
\begin{align}
    h_{t} = f_{\theta}(h_{t-1} + z_{steer}, x_{t})
    \label{eq: steering_vec_addition}
\end{align}
Here, we assume the dimension of~\zsteer~is equal to the model dimension $d^*$. If this is not the case, we can up or down project~\zsteer~without the addition of any parameters.
See~\citet{subramani2019can} for details.
We then optimize~\zsteer~to maximize the log-probability of the sequence as mentioned before via any off-the-shelf gradient-based optimization algorithm (\eg gradient descent, nonlinear conjugate descent, etc.).

To map steering vectors back to sequences (backward estimation), we intervene on a language model by injecting~\zsteer~and using beam search to decode starting with a \texttt{<bos>} token. We stop when an \texttt{<eos>} token or 100 total tokens is reached.
We measure how well the original sequence matches the generated sequence via three string overlap metrics: token-level exact match, BLEU score~\citep{papineni-etal-2002-bleu}, and longest prefix match.

\begin{figure*}[t]
    \centering
    \includegraphics[width=.95\linewidth]{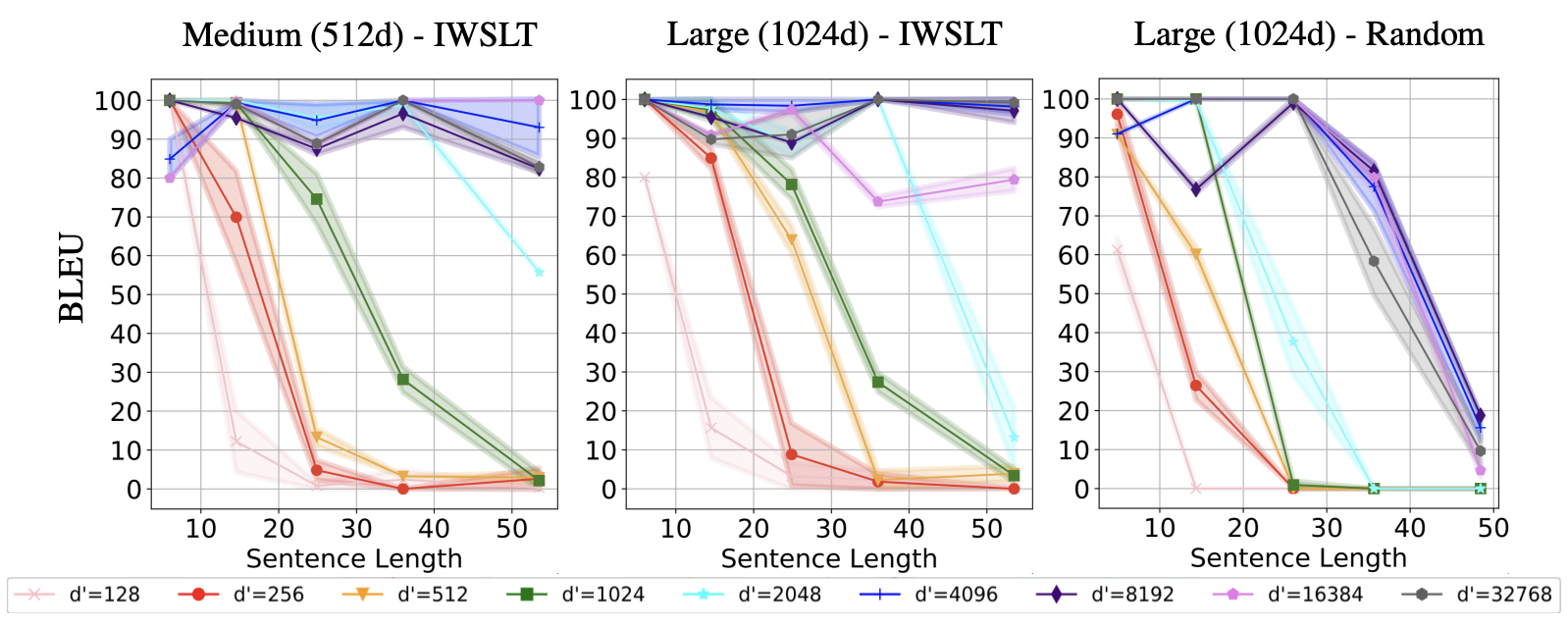}
    \caption{Recovery on IWSLT16 for \texttt{medium} (left) and \texttt{large} (center) and on random data for \texttt{large} (right).}
    \label{fig:recovery_iwslt_iwslt_random}
\end{figure*}

\paragraph{Experimental Setup}
First, we train our own language models on 50M sentences from the English Gigaword corpus~\citep{graff2003english}, with a 879k sentence development set and a 878k sentence test set stratified by article publishing date.
We use byte-pair encoding with 20,000 merges for a vocabulary of 20,234 subword tokens~\citep{Gage1994ANA, sennrich-etal-2016-neural}.
Our model is a 2-layer language model with LSTM units of three sizes, \texttt{small} ($d=256$), \texttt{medium} ($d=512$), and \texttt{large} ($d=1024)$ with shared input and output embeddings~\citep{press-wolf-2017-using}, and dropout~\citep{JMLR:v15:srivastava14a}.

To train the model, we use stochastic gradient descent with Adam with a learning rate of 1e-4 and a batch size of 100~\citep{kingma2014adam}. 
To learn steering vectors, we sample 100 randomly selected sentences from the development set as well as 50 sentences from the IWSLT16 En-De translation dataset to measure out-of-distribution generalization~\citep{cettolo-etal-2016-iwslt}.
We use nonlinear conjugate gradient~\citep{wright1999numerical} for optimization due to the highly non-convex nature of the objective function and use beam search with a width of 5 for backward estimation~\citep{graves2012sequence}.
See~\citet{subramani2019can} for more details.

\subsection{Takeaways}
We can find steering vectors for every sequence that achieve near perfect recoverability (token-level exact match $\geq$ 0.99) on \texttt{large}, offering an avenue for direct causal control.
Additionally, we find that larger, better trained models have higher recoverability and longer sequences are harder to recover.
One key question is whether our forward estimation procedure operates like a naive compressor without any structure.
In other words, does the forward estimation procedure have enough capacity to just encode the entire sequence in the vector without leveraging the language model's internal representations of language?
To test this, we create a \textit{random} dataset where we sample from the vocabulary with replacement at random where every token has equal probability.
We learn steering vectors for these sequences as well as the out-of-domain IWSLT16 data and measure recoverability.
In~\cref{fig:recovery_iwslt_iwslt_random}, we show that sequences that are lower entropy under the language model (IWSLT data) are much easier to recover at similar sequence lengths as compared to sequences from the \textit{random} data.
However, even for the very high entropy sequences (ones from \textit{random}),~\zsteer~has the capacity to encode sequences up to a token length of 28 nearly perfectly.

\paragraph{Limitations:}
Finding steering vectors was challenging. Optimization via conjugate gradient methods was very slow and could not be easily GPU-accelerated at the time, reducing adoption. 
Given how good recovery was for random sequences, a natural follow-up would be to understand what steering vectors encode and whether they could be useful beyond single sequence interventions.
There is no guarantee that the steering vector space learned here offers any additional utility.
This is precisely what we expand upon and explore in the next section.

\subsection{Bigger Picture}
At the time, being able to intervene on a language model with a single vector and causally force the model to generate \emph{any} sequence of interest without updating a single parameter was highly surprising.
This meant that language models had tremendous potential as universal decoders and steering could open up avenues to move away from task-specific finetuning and replace with inference-time steering.
Our work could serve as justification to attempt natural language prompting on better trained, stronger models, which occurred in the years that followed.
BERT had just recently come out~\citep{devlin-etal-2019-bert}, and while this paper was under review, we learned that BERT rediscovered the classical NLP pipeline~\citep{tenney-etal-2019-bert}, hinting that internal structure likely exists in transformer-based language models.

\section{Control: Steering Vectors for Transformers~\citep{subramani-etal-2022-extracting}}
\label{sec: steering_vectors_transformers}

We expand upon control, generalizing steering vectors to transformer-based language models for both \textit{exact steering} and \textit{concept-based steering}. We answer the following questions:
\begin{keyquestion}{Key Question 2}
Can transformer-based language models be steered to generate a desired sequence exactly without any parameter updates?
\end{keyquestion}
\begin{keyquestion}{Key Question 3}
Can extracted steering vectors act as useful representations with which we perform concept-based steering at inference-time?
\end{keyquestion}

\subsection{Prior Work}
Transformer language models started becoming popular, outperforming and largely replacing recurrent models~\citep{devlin-etal-2019-bert, radford2019language, Raffel2019ExploringTL}.
In~\cref{sec: steering_vectors_lstms}, we showed that LSTM-based LMs could be precisely controlled for short sequences with steering vectors, opening up the potential for them to be used as universal or general-purpose decoders.
Here, we explore whether higher-quality transformer-based language models could be more easily and efficiently steered, and thus make better universal decoder candidates. 
This work began prior to the release of GPT3~\citep{brown2020language}, hence the focus on small transformer-based models rather than LLMs.

\begin{table}[t]\small \centering
\begin{tabular}{@{}c|c|c@{}}
\toprule
\textbf{Injection location} & \textbf{Timestep} & \textbf{BLEU-4} \\ \midrule
Embedding & all timesteps & 33.99 \\ \midrule
\textbf{Layer 6 (self attn)} & \textbf{all timesteps} & \textbf{100.0} \\ \midrule
\textbf{Layer 6 (self attn)} & \textbf{first timestep} & \textbf{99.80} \\ \midrule
\textbf{Layer 7 (feed fwd)} & \textbf{all timesteps} & \textbf{100.0} \\ \midrule
\textbf{Layer 7 (feed fwd)} & \textbf{first timestep} & \textbf{99.25} \\ \midrule
\textbf{\makecell{All layers \\ (self attn + feed fwd)}} & \textbf{all timesteps} & \textbf{100.0} \\ \midrule
\textbf{\makecell{All layers \\ (self attn + feed fwd)}} & first timestep & 91.72 \\ \midrule
LM head & all timesteps & 6.72 \\ \bottomrule
\end{tabular}\caption{Sentence recovery for steering vectors when injected into different layers of the transformer model  and at multiple timesteps.}
\label{tbl:main_extraction}
\end{table}

\subsection{Our Contributions}
We coin the term \textit{steering vector}.
A vector~\zsteer~$\in \mathbb{R}^d$ is a \textit{steering vector} for a sequence $x$ under a model $M$ only if $M$ exactly generates $x$ via greedy decoding when~\zsteer~is injected into $M$.\footnote{Note that steering vectors need not correspond to an exact sequence. They are commonly now used to steer towards a desired concept or attribute.}

\paragraph{Background}
We define a flat-vector space $\mathcal{Z} \in R^{d'}$ for a transformer language model, similar to the recurrent language model from~\cref{sec: steering_vectors_lstms}.
To map the trajectory of hidden states for a sequence $x_1, \ldots, x_T$, we add a bias term~\zsteer~$\in \mathcal{Z}$ to the first-timestep at a single layer in the transformer stack after the feed-forward layer, see~\cref{fig:steering_vec_process_fig} for details.\footnote{\zsteer~could be added anywhere in the transformer stack and repeated across layers or timesteps, but we found that adding it once at a single layer and first timestep was sufficient.}
We also optimize~\zsteer~to maximize the log-probability of a given sequence, giving us the ability to map sequences to steering vectors (and vice-versa) and measure recoverability, exactly like in LSTM-based models.
This process is more efficient in transformers as compared to LSTMs because~\zsteer~is only added at one layer and one timestep.
We can up or down project~\zsteer~if we want to control the capacity of the steering vector.

\paragraph{Experimental Setup}
We take the GPT2-117M model and learn steering vectors by sampling sequences from four different genres (movies, books, news, and wikipedia) and stratify them based on length for a total of 256 sequences.
Recoverability is measured via BLEU score.
We vary where (injection location) and when (injection timestep) to intervene with~\zsteer.
For optimization during forward estimation, we use Adam with a learning rate of 1.0 and use greedy decoding to recover sequences.
We measure the extent to which steering vectors can be used as representations and compare them with mean-pooled hidden states.

Lastly, we explore whether concept-based steering is possible.
We first extract steering vectors for sequences for positive and negative sentiment respectively via the Yelp sentiment dataset~\citep{Shen2017StyleTF}.
We propose difference-of-means (DiffMean) steering, which works as follows.
First, we use vector arithmetic to take the mean of the positive sentiment steering vectors $z_{pos}$ and the negative sentiment ones $z_{neg}$.
This is then operationalized at inference-time.
For example, to steer towards positive sentiment, you add a steering vector $z_{steer} = \lambda(z_{pos} - z_{neg})$ at the timestep and location that those steering vectors were extracted from.\footnote{$\lambda \in \mathbb{R}$ is a constant known as the steering strength.}

\begin{table}[t]
\small
\centering
\begin{tabular}{c l}
\toprule
Positive Input & the taste is excellent!\\ \midrule 
+$0.5*(z_{neg} - z_{pos})$ & the taste is excellent!\\ 
+$1.0*(z_{neg} - z_{pos})$ & the taste is excellent!\\ 
+$1.5*(z_{neg} - z_{pos})$ & \makecell[l]{the taste is bitter and bitter\\taste is bitter taste is bitter} \\
\midrule
Negative Input & the desserts were very bland.\\ \midrule 
+$0.5*(z_{pos} - z_{neg})$ & the desserts were very bland.\\ 
+$1.0*(z_{pos} - z_{neg})$ & the desserts were very bland.\\ 
+$1.5*(z_{pos} - z_{neg})$ & the desserts were very tasty.\\ 
+$2.0*(z_{pos} - z_{neg})$ & the desserts were very tasty.\\ \bottomrule
\end{tabular}\caption{Concept steering for sentiment for a positive input sentence (top) and negative input sentence (bottom).}\label{tbl: sent-transfer-examples}
\end{table}

\subsection{Takeaways}
Our experiments show that fine-grained control via the exact steering of transformer-based language models is much easier and more efficient than the exact steering of LSTMs.
\cref{tbl:main_extraction} shows that nearly all sequences are perfectly recovered, even when adding~\zsteer~at just the first timestep.
As long as~\zsteer~is injected in the transformer stack (after the embedding and before the final layer), recoverability remains nearly perfect.
Linearly interpolating between steering vectors gives us a glimpse into what the steering vector space looks like.
Decoding from these intermediate points reveals structure: the space seems relatively smooth with large clusters corresponding to each of the sequences being interpolated between and a smooth transition in both syntax and semantics when moving from one sequence to another.
Cosine distances between steering vectors at middle layers reflect semantic similarity better than mean-pooled hidden states when measured on the semantic textual similarity benchmark~\citep{cer-etal-2017-semeval}, indicating that steering vectors may be better representations than the ones learned by the underlying language models.

Steering vectors provide coarse-grained control, too.
Our experiments on unsupervised sentiment transfer via DiffMean steering on the Yelp sentiment dataset show that a single direction in latent space learned via these steering vectors can flip sentiment reliably.
We show two examples in~\cref{tbl: sent-transfer-examples}.
For the first time, we show that concept steering at inference-time is possible.


\subsection{Bigger Picture}
As language model quality started improving, control became an achievable goal.
Two months after starting this project, GPT3 came out showing that large pretrained language models had the ability to, at inference-time, be few-shot prompted to solve different tasks.
Our work could serve as further justification for more ambitious inference-time based control such as in-context learning, alignment, and persona-based steering.
Over the past 5 years, language models became more performant with higher quality representations and concept-based steering took off, operating on the activation space rather than a reparametrized vector space like our steering vector space for inference-time control~\citep[\emph{inter alia}]{turner2023steering, li2023inference, Dunefsky2025OneshotOS, arad-etal-2025-saes, bigelow2026beliefdynamicsrevealdual, Morgulis2026SubliminalSS, wurgaft2026manifoldsteeringrevealsshared}.
\section{Trust: \textsc{mice}~\citep{subramani-etal-2025-mice}}
\label{sec: trust_mice}

We tackle trust by leveraging model internals to try to answer a key question: 
\begin{keyquestion}{Key Question 4}
Can we harness the latent spaces of language models to build better confidence estimators for tool-calling agents?
\end{keyquestion}

\begin{figure*}[t]
    \centering
    \includegraphics[width=0.55\textwidth]{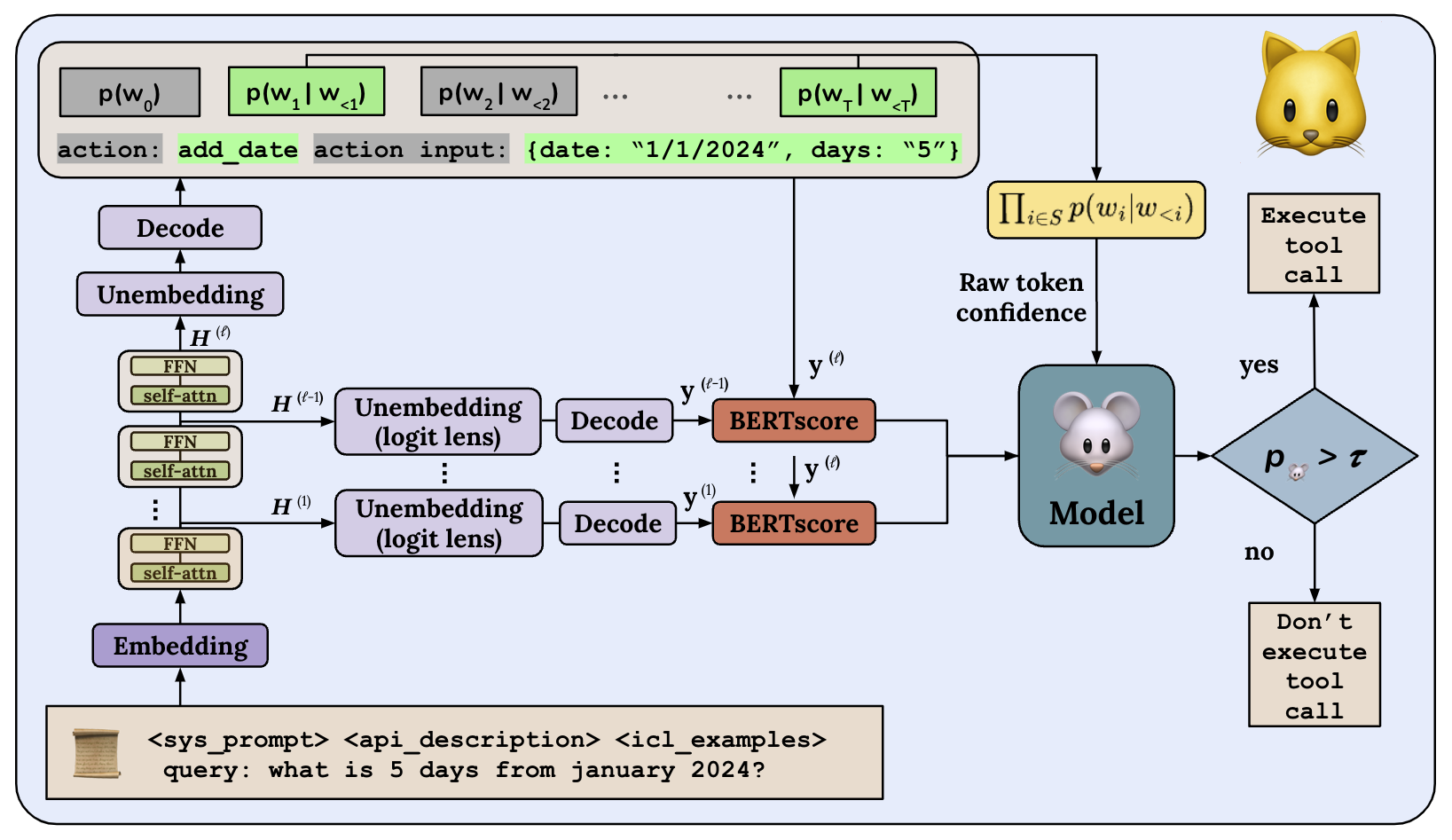}
    \hfill
    \includegraphics[width=0.43\textwidth]{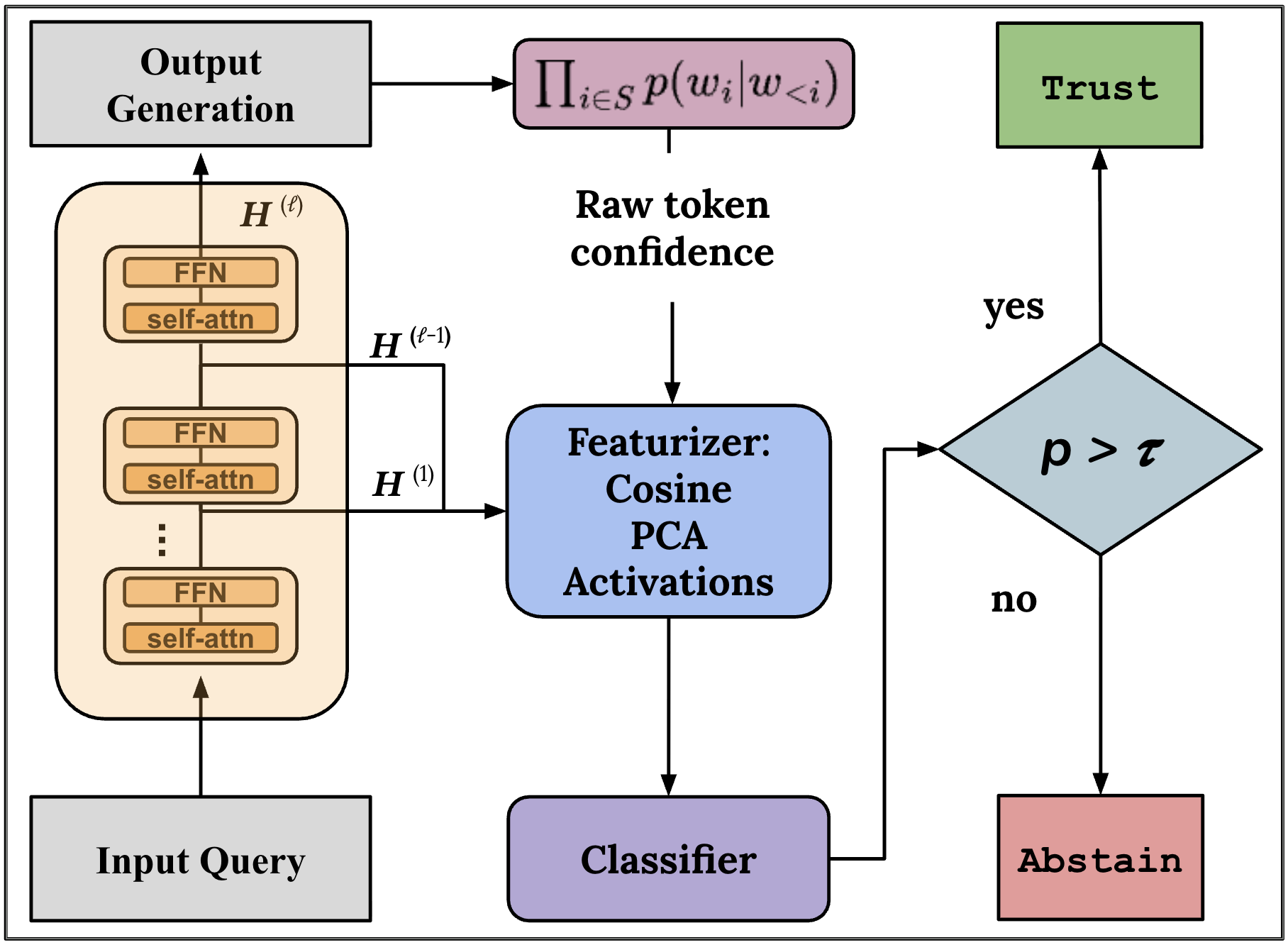}
    
    \caption{Here we show both the \mice~(left) and \acuteprotocol~(right) systems to calibrate language model generations. The only major differences between the systems are the featurizers and classifiers used.}
    \label{fig:mice_acute_process_fig}
\end{figure*}

\subsection{Prior Work}
A confidence estimator is a model that estimates the probability that a different model's output is correct.
Since language models have an internal confidence for its output already (\ie the joint probability of the generated sequence), auxiliary confidence estimators are rarely used.
However, raw confidences of language models are known to be poorly calibrated~\citep{desai-durrett-2020-calibration, jiang-etal-2021-know, zhong-etal-2023-non}.
To be well-calibrated, a confidence estimator must be correct approximately as often as it thinks it is~\citep{dawid1982well}.\footnote{Calibration is commonly measured using expected calibration error (ECE;~\citet{murphy1967verification, naeini2015obtaining}) and Brier Score~\citep{glenn1950verification}.}

Confidence estimation in NLP has been studied in tasks such as machine translation~\citep{niculescu2005predicting}, semantic parsing~\citep{stengel-eskin-van-durme-2023-calibrated}, and long-form text generation~\citep{band2024linguistic}.
Calibrating binary classifiers with a single input feature, the raw confidence, is common practice in machine learning with Platt scaling~\citep{platt1999probabilistic}, isotonic regression~\citep{barlow1972statistical}, beta calibration~\citep{pmlr-v54-kull17a}, and histogram regression estimators with adaptive binning~\citep{nobel1996histogram}.

Our angle, harnessing the latent spaces of models to build confidence estimation mechanisms, is unique.
In fact, there are very few studies that even combine mechanistic interpretability with confidence estimation or trust.
\citet{beigi-etal-2024-internalinspector} improves trustworthiness by using contrastive learning on activations and~\citet{liu2025llm} predicts correctness in question-answering tasks using probes learned on activations.

\begin{figure*}[t]
\centering
    \includegraphics[width=\linewidth]{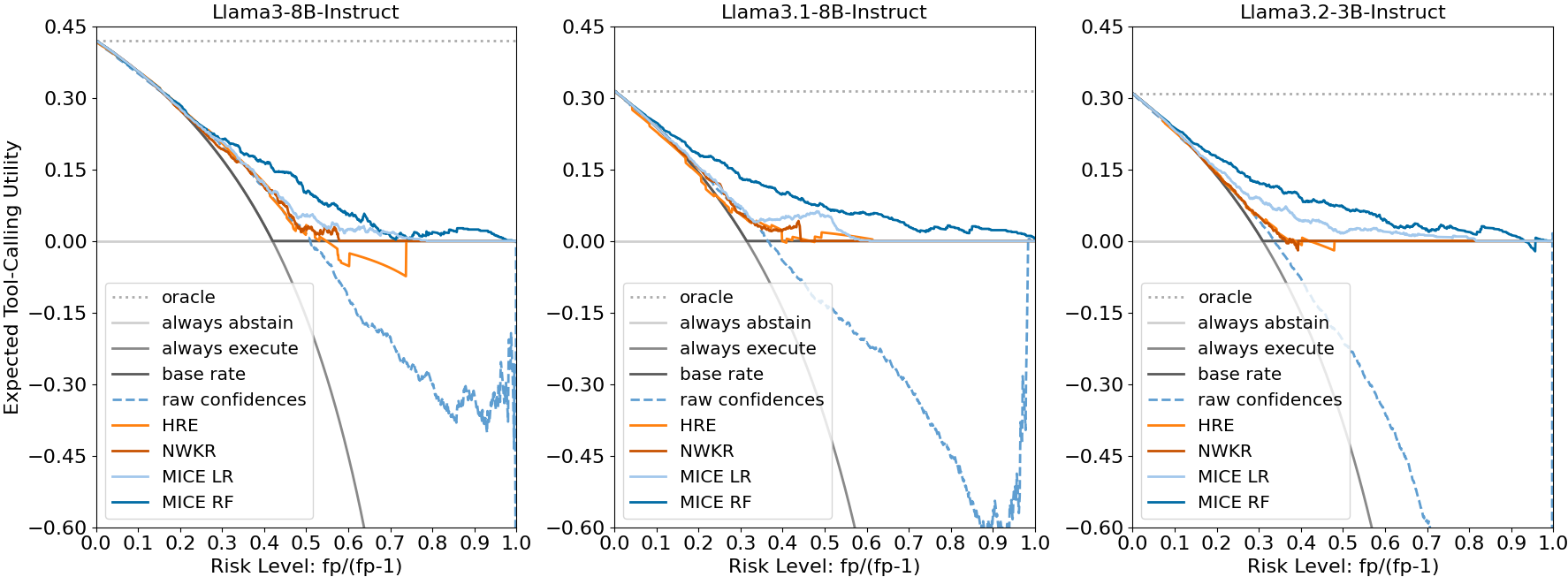}
    \caption{~\mice~systems outperform baselines on ETCU on the STE test set, especially at high-risk levels.}
    \label{fig:mice_results}
\end{figure*}

\subsection{Our Contributions}
\paragraph{Motivation:}
We explore whether we can leverage model internals to build a class of model internal confidence estimators (\mice) to better calibrate tool-calling agents.
Taking inspiration from the early-exiting and intermediate layer decoding literature (\ie the observation that a language model can often be decoded from early layers;~\citealp{geva-etal-2022-transformer, schuster2022confident, Belrose2023ElicitingLP, elhoushi-etal-2024-layerskip, yom-din-etal-2024-jump, merullo-etal-2024-language}), we theorize that a prediction that is slowly refined through the layers ought to be more trustworthy than one that suddenly appears in the final layer.

\paragraph{\mice:}
\cref{fig:mice_acute_process_fig} shows \mice~on the left.
For a given input query $q$, we pass it through the language model and at each layer $i$, we use \textit{logit lens} to decode from that intermediate layer~\citep{logit_lens}, resulting in a candidate generation $y^{(i)}$. 
Then, we compare how close that intermediate generation is to the final output generation via BERTscore (BERTscore($y^{(i)}, y^{(final)}$);~\citealp{zhang2019bertscore}) using DeBERTa-xlarge-mnli~\citep{He2020DeBERTaDB}, using those as features to train a correctness classifier. 
The probability that the classifier assigns to \texttt{<correct>} \emph{is} the new calibrated confidence.
In practice, we use both BERTscore features and the raw confidence (\ie the joint probability that the language model assigns to the output sequence) as features to train the classifier.\footnote{We experiment with two classifiers: a random forest and a logistic regressor.}

\paragraph{Measuring Trust:}
We care about evaluating how \textit{good} a confidence estimator is.
This is commonly measured using ECE. 
However, ECE suffers from two major drawbacks:
\begin{enumerate}[nolistsep]
\item Cannot distinguish between an oracle and a base rate estimator (\ie one that just predicts the base rate regardless of correctness)
\item Invariant to the risk level of the task
\end{enumerate}
These drawbacks hinder decision-making utility and thus calibration is necessary but not sufficient for our purpose.
We develop our own metric, expected tool-calling utility (ETCU) to solve these drawbacks, offering a better metric with which to evaluate the quality of a confidence estimator.\footnote{See~\citet{subramani-etal-2025-mice} for details.}
Additionally, we measure calibration error using a recently improved ECE variant called smooth ECE (smECE;~\citet{blasiok2024smooth}).

\paragraph{Experimental Details:}
Our experiments use the simulated trial-and-error (STE) dataset, a synthetically generated tool-calling dataset consisting of English-language queries, which call 50 distinct APIs~\citep{wang-etal-2024-llms-imaginarium}.
We split the data into demonstration (used to construct few-shot examples), training, validation, and test sets consisting of 4250, 1500, 750, and 750 examples each.
We 8-shot prompt three LLMs, Llama-3-8B-Instruct, Llama3.1-8B-Instruct, and Llama3.2-3B-Instruct and decode using greedy decoding to generate candidates~\citep{grattafiori2024llama3herdmodels}.
Our~\mice~classifiers are trained using the training set, hyperparameters are tuned using the validation set, and performance is evaluated on the test set.\footnote{A candidate generation is deemed to be correct if and only if it exactly matches the ground-truth answer.}

\subsection{Takeaways}
Our experiments confirm that language models are poorly calibrated. 
Traditional post-hoc recalibration techniques such as histogram regression (HRE) and Nadaraya-Watson kernel regression (NWKR) tend to be highly conservative~\citep{nadaraya1964estimating, watson1964smooth}, collapsing to base rate estimators with low calibration error, but low decision-making utility.
\mice~estimators, on the other hand, perform better, maintaining low calibration error, but higher decision-making utility due to a larger spread of probability estimates across examples.
In~\cref{fig:mice_results}, we observe that for tasks with medium or high-risk, both HRE and NWKR remain conservative, always abstaining for all inputs.
~\mice~performs better, correctly increasing its abstention rate as risk-levels increase, while trusting tool-calls some of the time.
To quantify how well a confidence estimator does across risk-level settings, we develop an area-under-the-curve (AUC) metric called \textsc{auc-etcu}, following~\citet{marcum1960statistical}.~\footnote{AUC style metrics are used in many areas of science~\citep[\emph{inter alia}]{Wagner1977BioavailabilityAM, geifman2019biasreduceduncertaintyestimationdeep,subramani-etal-2025-simba}.}

To test out-of-domain generalization for tool-calling, we create a scenario in which new APIs are tested on.
We hold out each of the 50 distinct APIs sequentially, resembling 50-fold cross validation and combine predictions across the entire test set.
Despite being zero-shot,~\mice~performs comparably to post-hoc calibration baselines across both smECE and ETCU, while having a larger spread of probability estimates across samples.

\paragraph{Limitations:}
~\mice~relies on an auxiliary model to calculate BERTscore features, which can be very expensive.
Additionally, we experiment with a single model family, Llama, on a single task, tool-calling.
Lastly, ETCU makes one strong simplifying assumption, that a confidence estimator should get a reward of 0 for abstaining regardless of whether the candidate generation was correct or not.
We address all of these limitations in~\cref{sec: trust_acute_euro}.

\section{Trust: \textsc{acute}~\citep{SubramaniAcute2026}}
\label{sec: trust_acute_euro}
We improve trustworthiness by addressing some limitations of~\mice~by asking: 
\begin{keyquestion}{Key Question 5}
Can we harness the latent spaces of language models to efficiently build better confidence estimators for model generations across a variety of tasks including multiple-choice question answering, tool-calling, and scientific document summarization?
\end{keyquestion}

\begin{table*}[t]
\footnotesize
\centering
\newcolumntype{d}{r@{.}l}
\newcommand\mc[1]{\multicolumn{1}{c}{#1}}
{\setlength{\tabcolsep}{3.5pt}
\begin{tabular}{l@{} c c c c c | c c c c c | c c c c c}
\toprule
& \multicolumn{5}{c}{\textbf{MMLU}} & \multicolumn{5}{c}{\textbf{APIGen}} & \multicolumn{5}{c}{\textbf{SCITLDR}}\\
\midrule
& ($\downarrow$) & \multicolumn{4}{c}{\textbf{\euroauc} ($\uparrow$)} & ($\downarrow$) & \multicolumn{4}{c} {\textbf{\euroauc} ($\uparrow$)} & ($\downarrow$) & \multicolumn{4}{c}{\textbf{\euroauc} ($\uparrow$)}\\
\cmidrule(r){3-6}
\cmidrule(r){8-11}
\cmidrule(r){13-16}
\textbf{estimator} & {\textbf{smECE}} & {low} & {med} & {high} & {all} & {\textbf{smECE}} & {low} & {med} & {high} & {all} & {\textbf{smECE}} & {low} & {med} & {high} & {all}\\
\midrule
{Raw Conf} & 0.17          & \underline{0.90} & 0.71 & 0.54 & 0.72 & 0.22 & 0.28 & 0.53 & 0.80 & 0.53 & 0.15 & 0.36 & \underline{0.71} & 	\textbf{0.92} & 0.66\\
{HRE}            & 0.11          & 0.87 & 0.72 & 0.71 & 0.77 & \textbf{0.02} & 0.82 & 0.70 & 0.82 & 0.78 & \textbf{0.08} & 0.67 & \underline{0.71} & \textbf{0.92} & \underline{0.77} \\
{\kernelReg}     & \textbf{0.07} & \underline{0.90} & 0.73 & 0.74 & 0.79 & \textbf{0.02} & 0.82 & 0.69 & 0.82 & 0.78 & \textbf{0.08} & 0.67 & \underline{0.71} & \textbf{0.92} & \underline{0.77} \\
\midrule
{\methodNameAbbr early act}  & \textbf{0.07} & \textbf{0.91} & 0.73 & 0.74 & 0.79 & 0.05 & \underline{0.90} & \underline{0.82} & \underline{0.87} & 0.86 & \textbf{0.08} & \underline{0.69} & \textbf{0.72} & \textbf{0.92} & \textbf{0.78}\\
{\methodNameAbbr mid act}    & \textbf{0.07} & \textbf{0.91} & \underline{0.76} & \underline{0.78} & \underline{0.82} & 0.06 & \underline{0.90} & \textbf{0.84} & \textbf{0.88} & \underline{0.87} & \textbf{0.08} & \textbf{0.70} & \textbf{0.72} & \textbf{0.92} & \textbf{0.78}\\
{\methodNameAbbr late act}   & \textbf{0.07}    & \textbf{0.91} & \textbf{0.77} & \textbf{0.80} & \textbf{0.83} & 0.07 & \underline{0.90} & \textbf{0.84} & \textbf{0.88} & \underline{0.87} & \textbf{0.08} & \underline{0.69} & \textbf{0.72} & \textbf{0.92} & \underline{0.77}\\
{\methodNameAbbr cosine}     & 0.09             & \underline{0.90} & 0.75 & \underline{0.78} & 0.81 & \underline{0.03} & 0.87 & 0.77 & 0.84 & 0.83 & \textbf{0.08} & 0.68 & \underline{0.71} & \textbf{0.92} & \underline{0.77} \\
{\methodNameAbbr pca10}      & \underline{0.08} & \textbf{0.91} & \underline{0.76} & \underline{0.78} & \underline{0.82} & 0.04 & \underline{0.90} & \underline{0.82} & \underline{0.87} & \underline{0.87} & \underline{0.09} & \textbf{0.70} & \textbf{0.72} & \textbf{0.92} & \textbf{0.78} \\
{\methodNameAbbr pca20}      & \underline{0.08} & \textbf{0.91} & \underline{0.76} & 0.77 & 0.81 & 0.06 & \textbf{0.91} & \textbf{0.84} & \textbf{0.88} & \textbf{0.88} & \underline{0.09} & \textbf{0.70} & \textbf{0.72} & \textbf{0.92} & \textbf{0.78} \\
 
\bottomrule
\end{tabular}
}
\caption{%
Results on the MMLU test set averaged across all 57 subtasks (left), on the APIGen test subset (middle), and on the SCITLDR dev set (right). All results for all tasks are averaged across the 6 LLMs we test. Lower smECE is better, while higher \euroauc is better. \textbf{Bold}, \underline{underline} indicate the best and second best result respectively. 
}
\label{tab:acute_results}
\end{table*}
\subsection{Our Contributions}
\paragraph{\acuteprotocol:} We introduce an activation-based confidence, utility, and trust estimation protocol (\acuteprotocol) to appropriately assess the confidence of a language model output.
In~\cref{fig:mice_acute_process_fig}, we show how~\acuteprotocol~works.
First, we take the activations and pass them through a featurizer.
Those features are fed into a classifier to predict whether the output generation is correct or not, exactly like~\mice.
Finally, the probability that the classifier assigns to correct is the new recalibrated confidence.
\acuteprotocol~removes the reliance on the auxiliary BERTscore model for input featurization.

\paragraph{\euro:}
We improve upon the ETCU metric from~\cref{sec: trust_mice} by removing the assumption that abstaining should always get 0 reward by introducing a new general and easily interpretable metric called expected utility renormalized by the oracle (\euro) that balances decision-making utility with calibration.
\euro~does not make simplifying assumptions, uses a single degree-of-freedom (the normalized net utility of correctly abstaining $u_{ca}$), and is bounded between an oracle estimator (\euro=1) and anti-oracle estimator (\euro=0).
Calculating~\euro~at different $u_{ca}$ or risk values traces a curve.
Measuring the area under that curve provides a score with which to compare confidence estimators called \euroauc.\footnote{See~\citet{SubramaniAcute2026} for further details, including a detailed derivation in the Appendix of that paper.}

\paragraph{Experimental Setup}
We apply~\acuteprotocol~to six new LLMs on three new tasks: multiple-choice question answering (MMLU;~\citealp{Hendrycks2020MeasuringMM}), tool-calling (APIGen;~\citealp{liu2024apigen}), and scientific document summarization (SCITLDR;~\citealp{cachola-etal-2020-tldr}).\footnote{Results are averaged across LLMs.}
Performance is measured via smECE and~\euroauc.

\begin{figure*}[t]
\centering
    \includegraphics[width=0.95\linewidth]{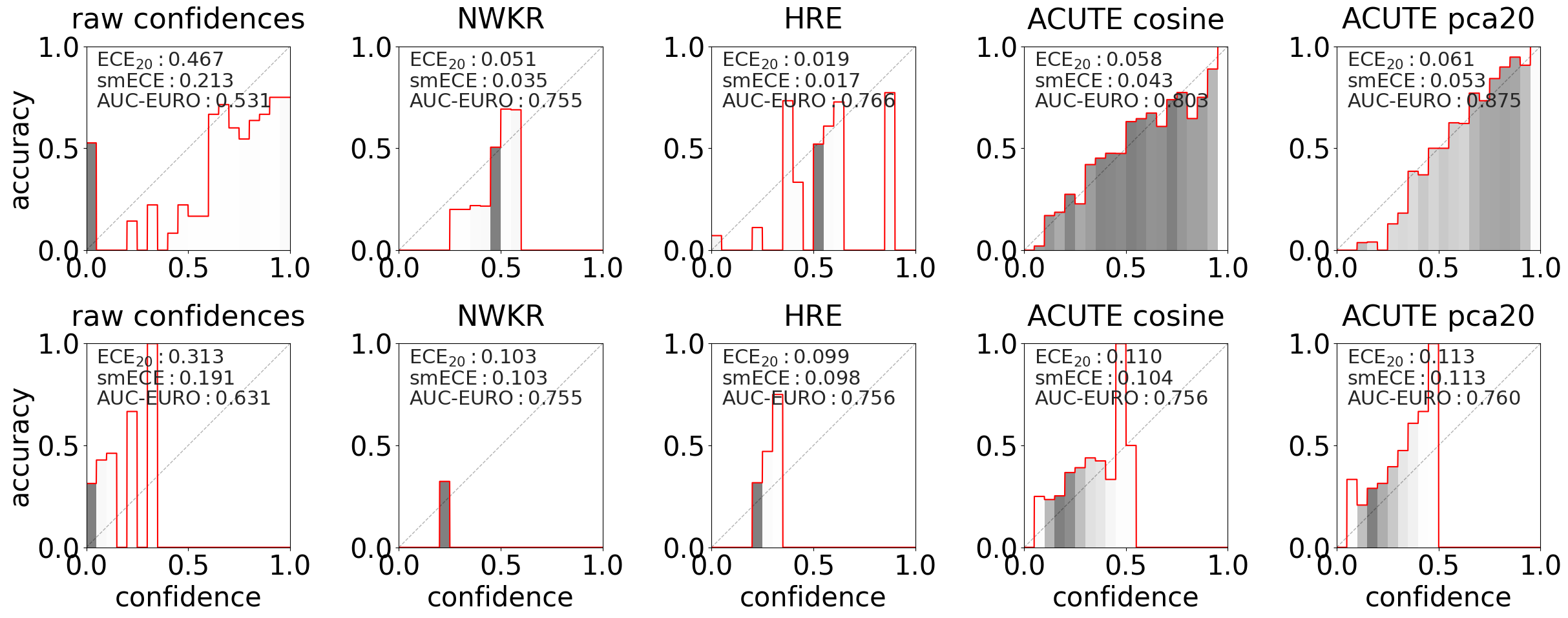}
    \caption{Reliability diagrams for the gemma-3-12b-it model for APIGen (top row) and SCITLDR (bottom row) across 3 baseline estimators and two~\acuteprotocol~ confidence estimators. Darker shading corresponds to higher density of examples in that confidence bin.}
    \label{fig: reliability_diagrams_gemma3}
\end{figure*}

\subsection{Takeaways}
\cref{tab:acute_results} and~\cref{fig: reliability_diagrams_gemma3} reveals that raw confidences remain poorly calibrated and post-hoc recalibration baselines collapse to base rate estimators as in~\cref{sec: trust_mice}.
\acuteprotocol~performs well across all three tasks, outperforming baselines on \euroauc, while maintaining low smECE.
Using activations from later layers (\acuteprotocol~late act) and PCA-reduction of all layer activations to 20 components (\acuteprotocol~pca20) performed best across all tasks.

\subsection{Bigger Picture}
Without any post-hoc calibration, language models provide terrible confidence estimates.
Most post-hoc calibration methods collapse the often large spread of probability estimates into a much narrower range, reducing calibration error without increasing decision-making utility.
Despite these drawbacks, probability estimates with or without post-hoc calibration remain the \textit{de-facto} standard.
Our work challenges this by proposing decision-making utility centric confidence estimators that can better adjudicate trust.
Further study on this front can help us develop even better confidence estimators and increase the reliability and safety of LLMs.
Overall, our work is a small step towards harnessing the latent spaces to appropriately assign trust to LLM outputs.
\section{Conclusion}
\label{sec: conclusion}
We present four research contributions that demonstrate how we can operationalize the latent spaces of language models for better control and trust.
Our work introduces steering vectors for exact and concept-based control on both LSTM- and transformer-based models.
Steering vectors provide nearly perfect fine-grained control at inference-time, suggesting that language models could be universal decoders.
We propose two methods which harness the latent spaces of models to learn confidence estimators for language model generations to improve trust.
Our methods recalibrate model outputs across architectures and tasks effectively, suggesting that the latent spaces contain information to build more reliable and trustworthy language technology, especially in high-stakes scenarios.
We hope that our work encourages others to open up the black box and study the latent spaces of language models.

\section*{Acknowledgments}
\label{sec: acknowledgements}
We thank the numerous collaborators and authors on each of the individual papers discussed here and both Mona Diab and Nivedita Suresh for feedback on an early version of this work.

\bibliography{all_bibtextidy_abbr}

\end{document}